\documentclass{article}

\PassOptionsToPackage{compress}{natbib}
\usepackage[final]{nips_2018}

\usepackage[utf8]{inputenc} %
\usepackage[T1]{fontenc}    %
\usepackage{hyperref}       %
\usepackage{url}            %
\usepackage{booktabs}       %
\usepackage{amsfonts}       %
\usepackage{nicefrac}       %
\usepackage{microtype}      %
\usepackage{times}  %
\usepackage{helvet}  %
\usepackage{courier}  %
\usepackage{url}  %
\usepackage{graphicx}  %
\usepackage{subcaption}
\usepackage{amsmath}
\usepackage{amssymb}
\usepackage{algorithm}
\newtheorem{thm}{Theorem}
\title{Can VAEs Generate Novel Examples?}
\usepackage{mathtools}
\DeclarePairedDelimiterX{\klx}[2]{(}{)}{%
  #1\;\delimsize\|\;#2%
}

\renewcommand{\vec}[1]{\ensuremath{\boldsymbol{#1}}}
\renewcommand{\v}[1]{\vec{#1}}

\newcommand{\x}{\ensuremath{\v{x}}}

\newcommand{\z}{\ensuremath{\v{z}}}

\newcommand{\p}{\ensuremath{p_\theta}}
\newcommand{\q}{\ensuremath{q_\phi}}

\newcommand{\E}{\ensuremath{\mathbb{E}}}

\author{
   Alican Bozkurt\\
   Northeastern University\\
   Boston, MA 02115\\
   \texttt{alican@ece.neu.edu}\\
   \And
   Babak Esmaeili\\
   Northeastern University\\
   Boston, MA 02115\\
   \texttt{esmaeili.b@husky.neu.edu}\\
   \And
   Dana H. Brooks\\
   Northeastern University\\
   Boston, MA 02115\\
   \texttt{brooks@ece.neu.edu}\\
   \And
   Jennifer G. Dy\\
   Northeastern University\\
   Boston, MA 02115\\
   \texttt{jdy@ece.neu.edu}\\
   \And
   Jan-Willem van de Meent\\
   Northeastern University\\
   Boston, MA 02115\\
   \texttt{j.vandemeent@northeastern.edu}\\
}

\begin{document}
\maketitle

\begin{abstract}
An implicit goal in works on deep generative models is that such models should be able to generate novel examples that were not previously seen in the training data. In this paper, we investigate to what extent this property holds for widely employed variational autoencoder (VAE) architectures. VAEs maximize a lower bound on the log marginal likelihood, which implies that they will in principle overfit the training data when provided with a sufficiently expressive decoder. In the limit of an infinite capacity decoder, the optimal generative model is a uniform mixture over the training data. More generally, an optimal decoder should output a weighted average over the examples in the training data, where the magnitude of the weights is determined by the proximity in the latent space. This leads to the hypothesis that, for a sufficiently high capacity encoder and decoder, the VAE decoder will perform nearest-neighbor matching according to the coordinates in the latent space. To test this hypothesis, we investigate generalization on the MNIST dataset. We consider both generalization to new examples of previously seen classes, and generalization to the classes that were withheld from the training set. In both cases, we find that reconstructions are closely approximated by nearest neighbors for higher-dimensional parameterizations. When generalizing to unseen classes however, lower-dimensional parameterizations offer a clear advantage.
\end{abstract}

\section{Introduction}
\noindent
Variational autoencoders \citep{kingma_auto-encoding_2013, rezende2014stochastic}  jointly train a generative model $\p(\x,\z)$ and an inference model $\q(\z,\x)$. The generative model is defined in terms of a likelihood $\p(\x \mid \z)$ and a prior $p(\z)$. The likelihood is parameterized using a neural network, which we refer to as the \emph{decoder}, where the prior is most commonly a spherical Gaussian. The inference model is defined in terms of a conditional $\q(\z \mid \x)$, parameterized by an \emph{encoder} network and the empirical distribution $\hat{p}(\x)$,
\begin{equation*}
    \mathcal{L}(\theta, \phi) 
    =
    \E_{\hat{p}(\x)}
    \left[
      \E_{\q(\z|\x)}
      \left[
      \log \frac{\p(\x, \z)}{\q(\z|\x)}
      \right]
    \right]
    \leq
    \E_{\hat{p}(\x)}
    \left[
    \log \p(\x)
    \right],
    \qquad
    \hat{p}(\x) 
    =
    \frac{1}{N}
    \sum_{n=1}^N
    \delta_{\x_n}(\x)
    .
\end{equation*}

One of the basic premises for works on deep generative models (including VAEs) is that we expect such models to not only reproduce the data that they are trained on faithfully, but also to generate entirely novel but plausible samples. The ambiguity in the terms ``novel'' and ``plausible'' in this premise induces a vague idea about generalization.

Even though discriminative models have been enjoying a clear definition of ``generalization performance'' and guaranteed generalization bounds due to statistical learning theory \citep{wang_identifying_2018}, works on generalization in generative models started to emerge only very recently, without a widely agreed metric, and honestly, without even a clear definition of generalization.

While generalization has been simply defined as the performance on a test dataset, here we concentrate on a more challenging task. Under the \emph{manifold hypothesis}, real-world data presented in high dimensional spaces are expected to lie in a manifold-space of much lower dimensionality \citep{bengio2013representation}. The data in the train and test sets can be different in the high-dimensional space, but identical in the feature space. We define generalization as the ability to reason about unseen data that is significantly different than the training data in the feature space. For example, we can imagine a dataset containing different shapes and colours such that all shapes and colours exist, but not all combinations exist. A model that is able to generalize well must perform well on combinations of shapes and colours not seen during training. 

An interesting observation made by several researchers \citep{bousquet_optimal_2017, rezende_taming_2018, alemi2018fixing}, but given in its most general from in \cite{shu_amortized_2018}, is that the reconstruction obtained from an optimal decoder of a VAE is a convex combination of examples in the training data.

\begin{thm}[\cite{shu_amortized_2018}]
Let $\mathcal{P}$ be an exponential family with corresponding mean parameter space $\mathcal{M}$ and sufficient statistic function $T(\cdot)$. Consider $\z \in \mathcal{Z}$, $g \in \mathcal{G} :\mathcal{Z}\rightarrow \mathcal{M}$, and a fixed $\q(\z|\x)$. Supposing $\mathcal{G}$ has infinite capacity, then the optimal generative model $g^*$ returns:
\begin{equation}
\mu = g^*(\z) = \sum_{i=1}^n\q(\x^{(i)}|\z)T(\x^{(i)})=\sum_{i=1}^n\frac{\q(\z|\x^{(i)})}{\sum_j\q(\z|\x^{(j)})}T(\x^{(i)})=\sum_{i=1}^nw_i(\z)T(\x^{(i)}).
\end{equation}
\label{shu-theorem}
\end{thm}
In the specific case of a Bernoulli likelihood, Theorem~\ref{shu-theorem} states that the reconstructed image is simply a weighted average over images observed during training, where the magnitude of the weights is governed by the proximity to training examples in the latent space. For the rest of this paper, when we talk about ``weighted average`` or ``weights'', we refer to the terms $\mu$ and $w_i$ in Theorem~\ref{shu-theorem} respectively, and we use $\hat{\x}$ to refer to the output of the decoder. 

Theorem~\ref{shu-theorem} raises a number of interesting questions. How closely do commonly trained VAEs mirror this behavior in practice? How strong is the infinite capacity assumption and what is the relationship between the model capacity and generalization? On average, what does the distribution of weights look like for a given test instance? 

To answer these questions, we perform a series of experiments, explained in the following section. In the heart of all our experiments lies this hypothesis: if Theorem~\ref{shu-theorem} holds, then VAEs should not able to reconstruct an out-of-training-distribution sample. 

\begin{figure}[htb]
    \centering
    \includegraphics[width=0.9\linewidth]{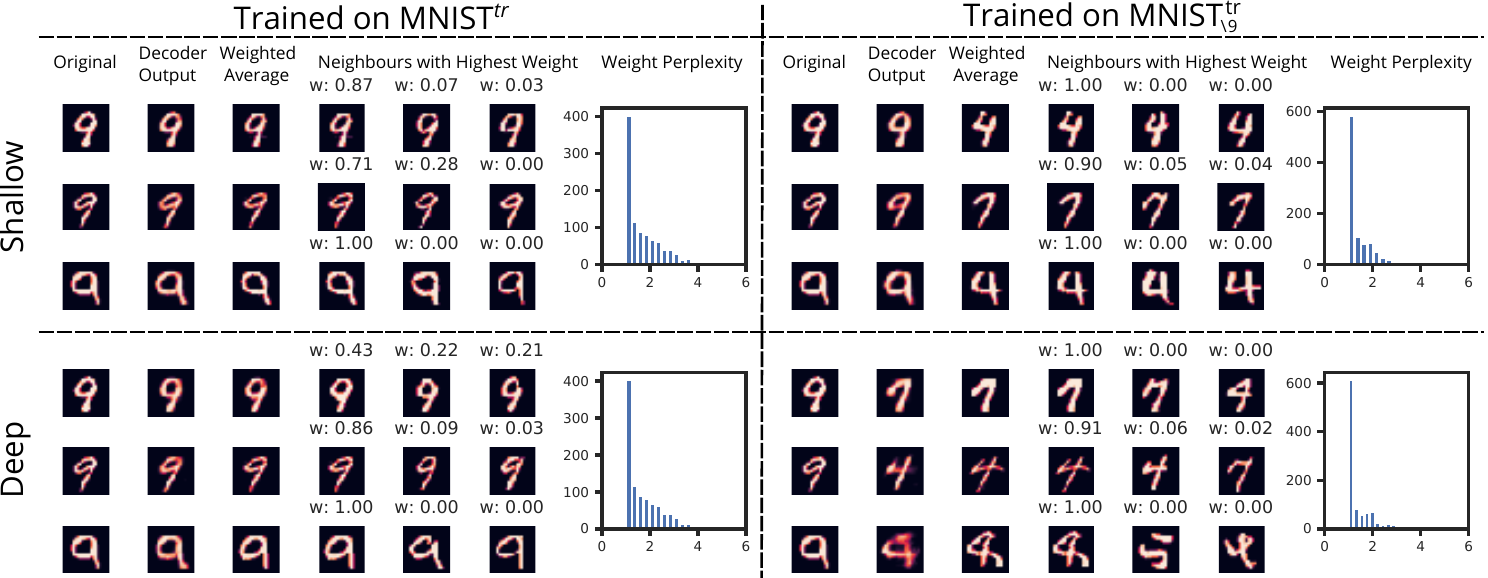}
    \caption{Reconstruction of seen vs unseen classes for shallow and deep VAE.}
    \label{fig:9-removal}
    \vspace{-1em}
\end{figure}

\section{Experiments}
We will be using different partitions of the MNIST dataset in our experiments, so we find it useful to introduce a notation for brevity. Let $\text{MNIST}^{te}_{d}$ be the dataset containing samples of digit $d$ from test set of MNIST. Similarly, $\text{MNIST}^{tr}_{\backslash d}$ denotes the dataset containing samples of all digits except $d$ from training set of MNIST. When used without any subscripts, e.g. $\text{MNIST}^{tr}$, it indicates samples from all digits are present in the dataset.   

In order to evaluate in-distribution and out-of-distribution generalization performance of VAEs, we perform the following experiment. We train a shallow and a deep VAE with $\text{dim}(\z)=50$ on two datasets based on MNIST: $\text{MNIST}^{tr}$ and $\text{MNIST}^{tr}_{\backslash 9}$. All networks are evaluated on the same test dataset $\text{MNIST}^{te}_{9}$. In the shallow encoder and decoder, we have used a single hidden layer (400 neurons), and their deep counterparts have three hidden layers (400, 200, and 100 neurons). After training, we evaluate all four models by feeding them samples from $\text{MNIST}^{te}_{9}$ and observing the decoder output, the weighted average calculated according to \eqref{shu-theorem}, and the three training examples with the highest weight (see Figure~\ref{fig:9-removal}). 

We make three observations here. First, for both VAEs trained on $\text{MNIST}^{tr}$, reconstruction and weighted average images look quite similar (see "Trained on $\text{MNIST}^{tr}$" column in Figure~\ref{fig:9-removal}). Second, for most test samples presented, it is only a single training example that contributes to the weighted average. We confirm this quantitatively by computing the perplexity histograms of the weights for all test data, which we show in Figure~\ref{fig:9-removal}. For all cases, the histograms have a high peak at 1, indicating that for most test examples, a single training sample has weight 1 and the rest have zero weight.\footnote{Note that the weights are normalized Gaussian likelihoods, so $w_i \in (0,1)$, but for weights close to zero, effect of the corresponding training sample is negligible.} Note that this is equivalent to $k$-nearest neighbour matching with $k=1$.

\begin{figure}[htb]
    \centering
    \begin{minipage}{.48\textwidth}
    \vspace{-0.5em}
    \includegraphics[width=\linewidth]{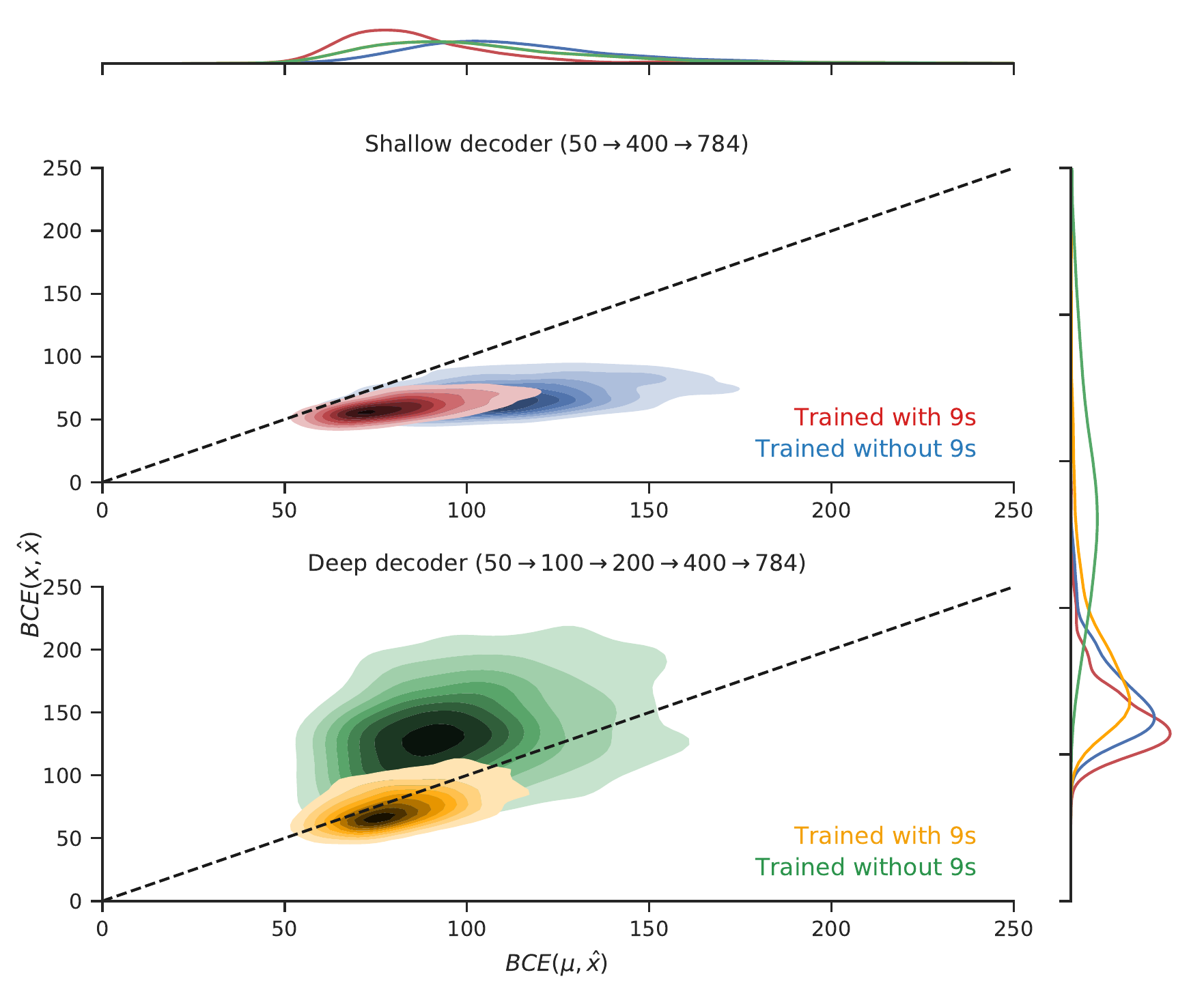}
    \vspace{-0.5em}
    \caption{Distribution of binary cross entropy loss between input image $\x$ and the decoder output $\hat{x}$, vs the loss between weighted average image $\mu$ and $\hat{x}$; calculated over $\text{MNIST}^{te}_9$.}
    \label{fig:bce}
    \end{minipage}
    \hspace{1ex}
    \begin{minipage}{.48\textwidth}
    \includegraphics[width=\linewidth]{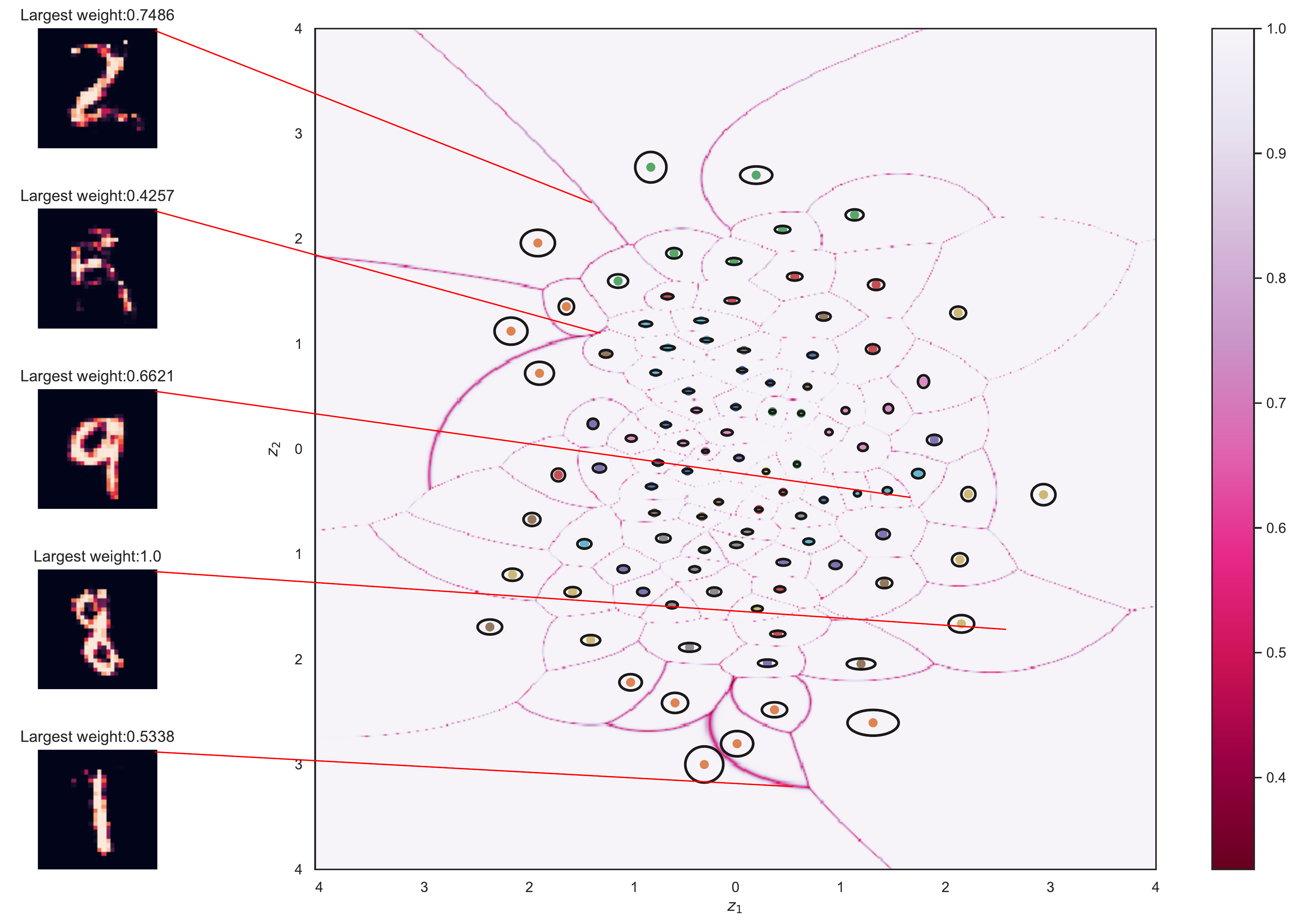}
    \caption{mini-MNIST latent space: points and ellipses show posterior distribution, and colormap shows the magnitude of largest weight. The images on the left are decoder outputs of the $\z$s chosen via stratified sampling, along with their weights.}
    \label{fig:voronoi}
    \end{minipage}%
\end{figure}

Perhaps the most fascinating finding here is the shallow VAE's ability to generalize to out-of-training-distribution samples: It can reconstruct perfectly passable nines even though it has not seen one during training. Furthermore, the assumption of infinite capacity in Theorem~\ref{shu-theorem} clearly matters as the decoder outputs are not similar to the weighted average for the shallow VAE trained on $\text{MNIST}^{tr}_{\backslash9}$. As shown in Figure~\ref{fig:9-removal}, the reconstructions are more similar to the input images in the shallow VAE, while being closer to the weighted average in the deep VAE. We confirm this by comparing the binary cross entropy loss of reconstructions to the samples from withheld class and the weighted average (see Figure~\ref{fig:bce}). The reconstructions are closer to the input than the weighted average for the shallow VAE (most of the mass of blue distribution lies below the line). On the other hand, the reconstructions are closer closer to the weighted average than the input for the deep VAE (most of the mass of green distribution lies above the line).  

One final counter-intuitive observation is that increasing the complexity affects encoder and decoder differently. For decoder, we see some change in characteristics of the reconstructions, whereas the weight perplexity histograms are similar for both deep and shallow cases, which can be considered as a characteristic of the encoder.

To visualize the latent space, we trained a VAE with $\text{dim}(\z)=2$. Since our infinite capacity assumption simply would not hold for $\text{MNIST}^{tr}$, we uniformly sample 10 samples per digit from $\text{MNIST}^{tr}$, and train our VAE on this ``mini-MNIST'' dataset. Figure~\ref{fig:voronoi} shows the latent space, where the points and ellipses show the posterior distribution, and the overlaid colormap shows the largest weight for a given $\z$ ($\max_i w_i(\z)$). First observation is posterior distributions are converging to delta distributions which is an indicator of optimality \citep{rezende_taming_2018}. Second, for most of the latent space, maximum weight is 1, i.e. the weighted average is the nearest neighbour in latent space.   

\subsection{Role of Encoder/Decoder Capacity in Generalization}

As with ``generalization'', ``capacity'' of a neural network is a hard aspect to characterize. Recent work has attempted to characterize capacity of discriminative models and datasets using algebraic topology~\citep{guss2018}, but there are no equivalent results for their generative counterparts. Here, we will use number of parameters and layers as simple proxies for capacity. 

\begin{figure}[htb]
    \vspace{-1em}
    \centering
    \includegraphics[width=0.9\linewidth]{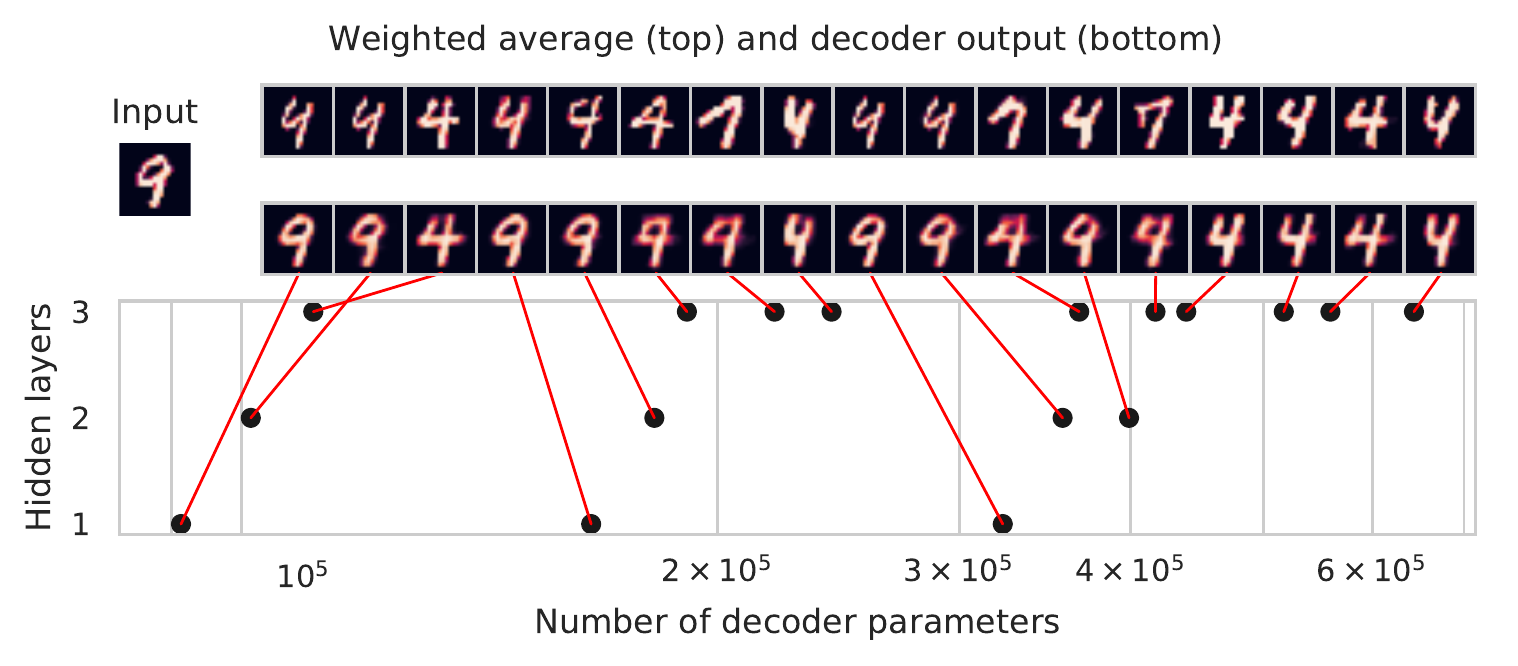}
    \vspace{-0.5em}
    \caption{Decoder outputs and weighted average images for VAEs with different architectures.}
    \label{fig:complexity}
    \vspace{-1.0em}
\end{figure}

In the previous experiments, we provided a comparison between a shallow and a deep VAE. Here, we do a finer grained complexity analysis, and identify regions where Theorem~\ref{shu-theorem} holds. In Figure~\ref{fig:complexity}, we show reconstructed and weighted average images for 17 VAEs with different network architectures given an input sample from the withheld class. As VAEs get more complex, they overfit the training data therefore fail to reconstruct the unseen digit. Moreover, we observe that the reconstruction is closer to the weighted average for higher capacity networks. Another -intuitive- observation here the number of layers plays a more crucial role in complexity than number of parameters, since for VAEs with 3 hidden layers, reconstructions are more similar to weighted average, while in single hidden layer VAEs, the reconstructions match the input sample regardless of the number of parameters.

\section{Conclusion}

The theorem by \citeauthor{shu_amortized_2018} provides an interesting viewpoint on how VAEs reason about unseen data. By investigating how much this theorem applies in practice, we uncovered some interesting properties about VAEs. In particular, we studied the connection between network capacity and generalization. Our findings show that networks with restricted capacity generalize better to out-of-training-distribution samples. For networks with sufficiently high capacity, we found that the number of training samples accountable for reconstructing a sample is often quite low, which indicates that generative capability of such VAEs are similar to a generator that employs nearest neighbour matching. We also found that VAEs with larger number of layers behave more consistently with Theorem~\ref{shu-theorem} in comparison to number of parameters. For future work, it may be helpful to investigate objectives where the optimal generative model combines different \emph{features} of training samples rather than averaging them in the high-dimensional space.

\bibliographystyle{plainnat}
\bibliography{vae-generalization}

\newpage
\appendix
\section{Complexity Diagrams for Different Held-out Digits}

\begin{figure}[htb]
    \vspace{-1em}
    \centering
    \includegraphics[width=0.9\linewidth]{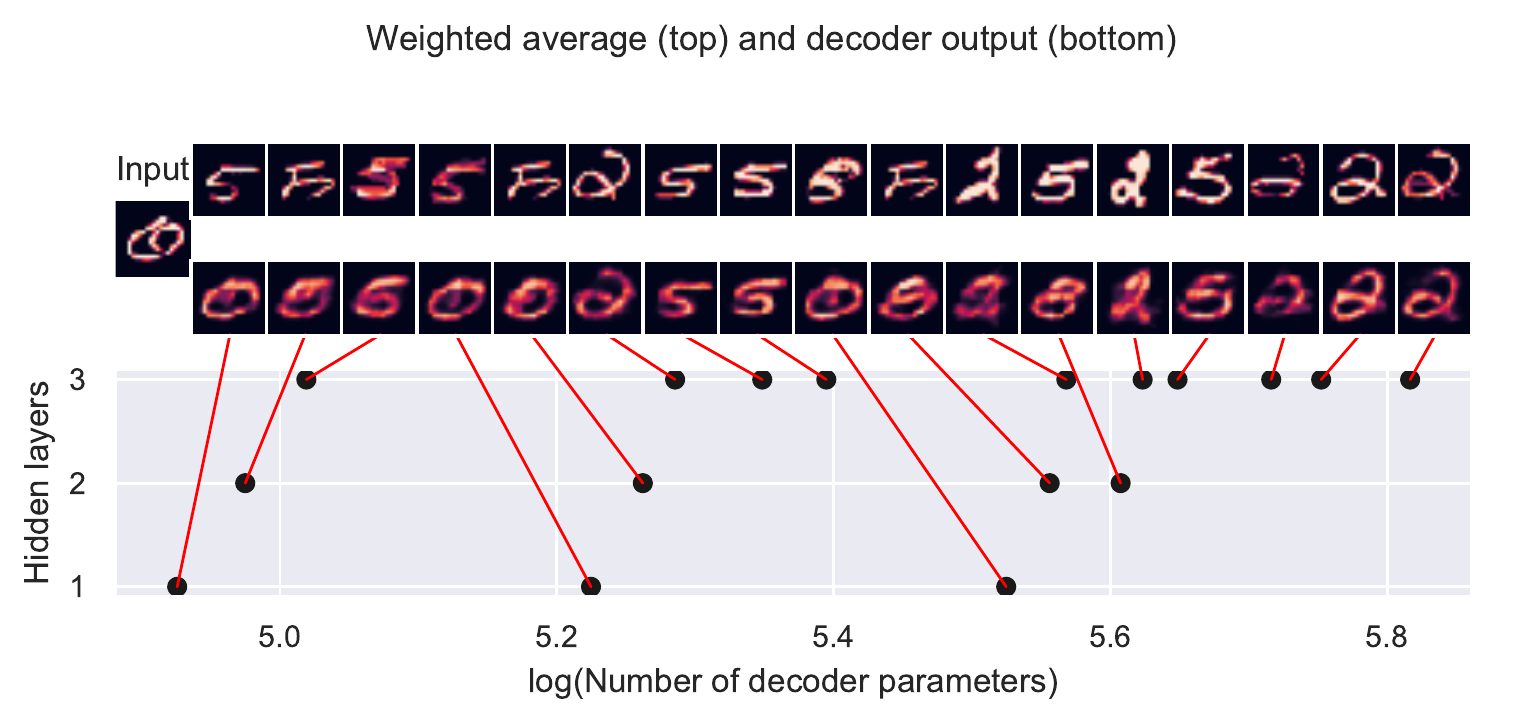}
    \vspace{-0.5em}
    \caption{Decoder outputs and weighted average images for VAEs with different architectures trained on $\text{MNIST}^{tr}_0$.}
    \vspace{-1.0em}
\end{figure}

\begin{figure}[htb]
    \vspace{-1em}
    \centering
    \includegraphics[width=0.9\linewidth]{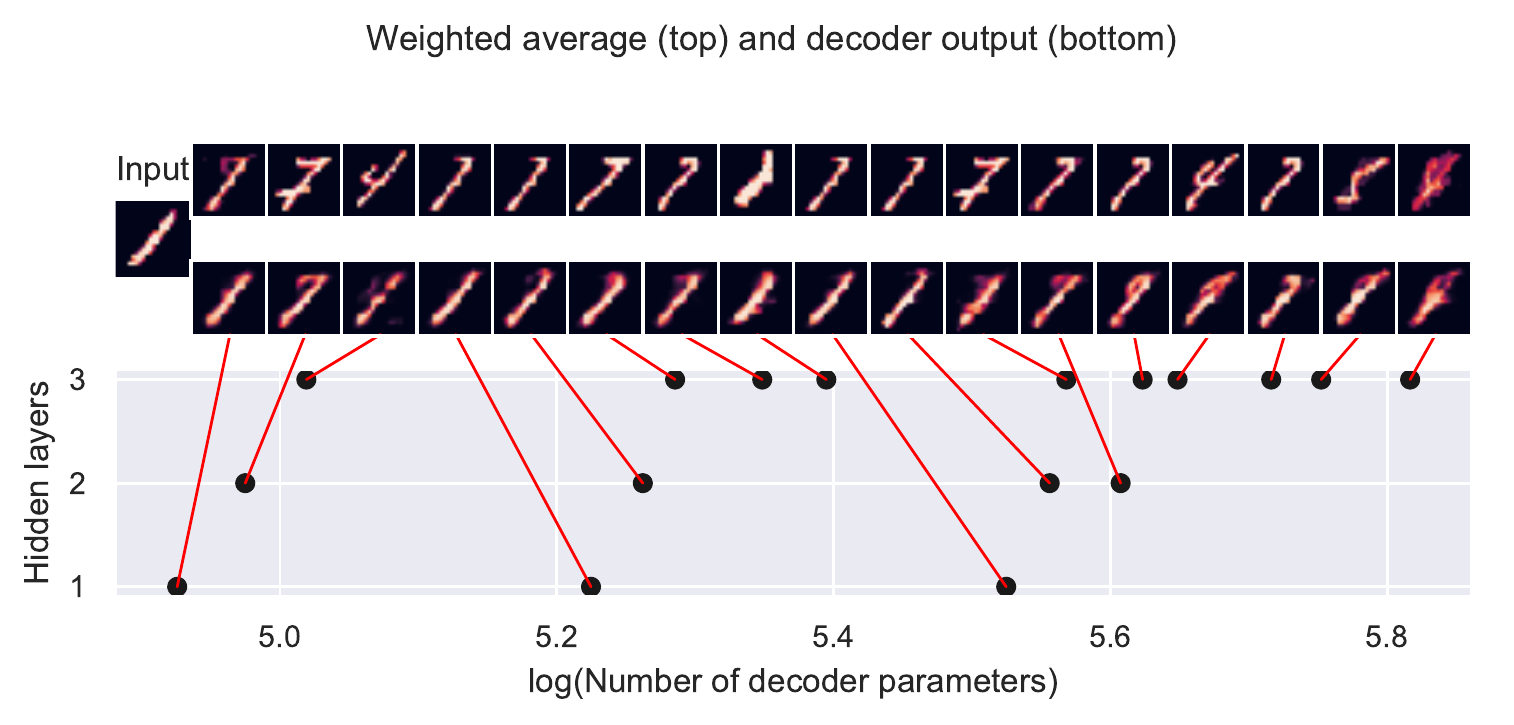}
    \vspace{-0.5em}
    \caption{Decoder outputs and weighted average images for VAEs with different architectures trained on $\text{MNIST}^{tr}_1$.}
    \vspace{-1.0em}
\end{figure}

\begin{figure}[htb]
    \vspace{-1em}
    \centering
    \includegraphics[width=0.9\linewidth]{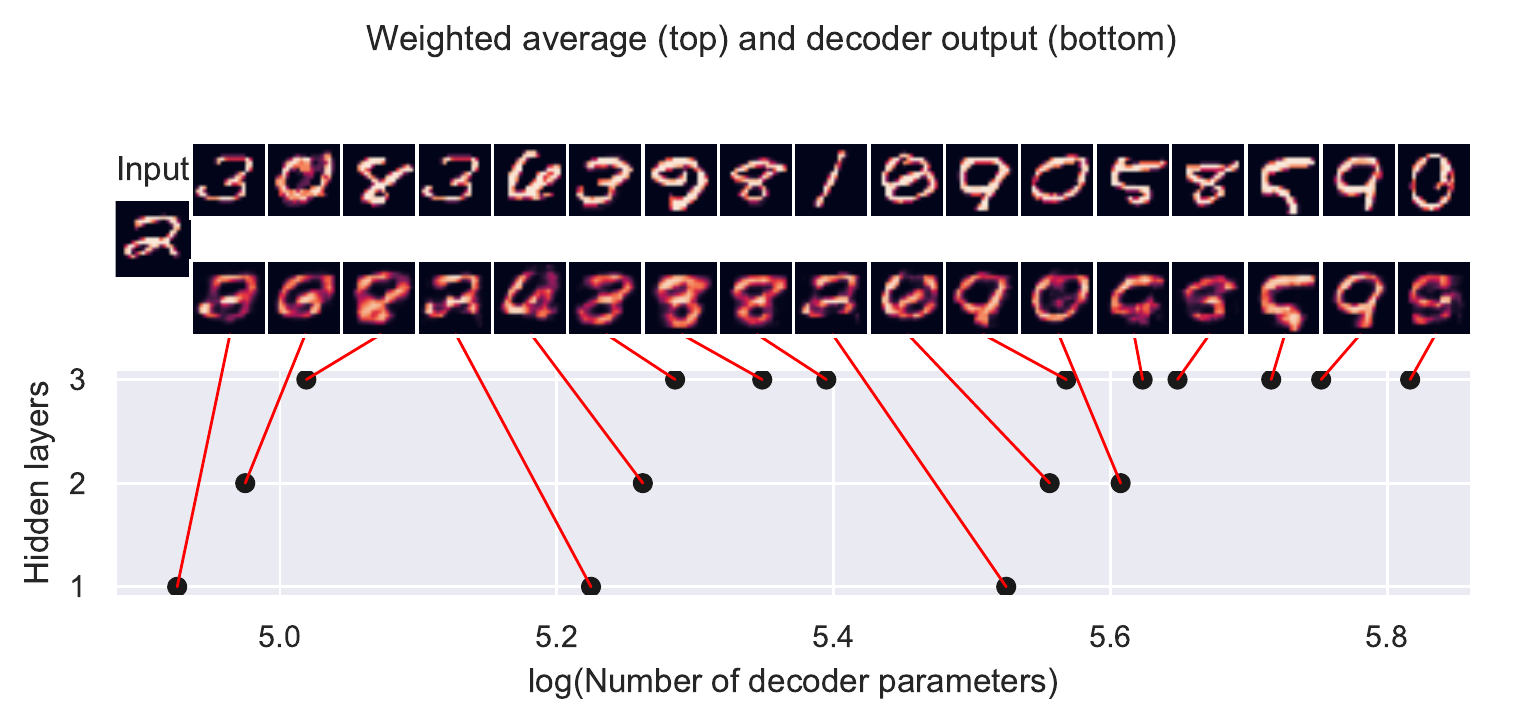}
    \vspace{-0.5em}
    \caption{Decoder outputs and weighted average images for VAEs with different architectures trained on $\text{MNIST}^{tr}_2$.}
    \vspace{-1.0em}
\end{figure}

\begin{figure}[htb]
    \vspace{-1em}
    \centering
    \includegraphics[width=0.9\linewidth]{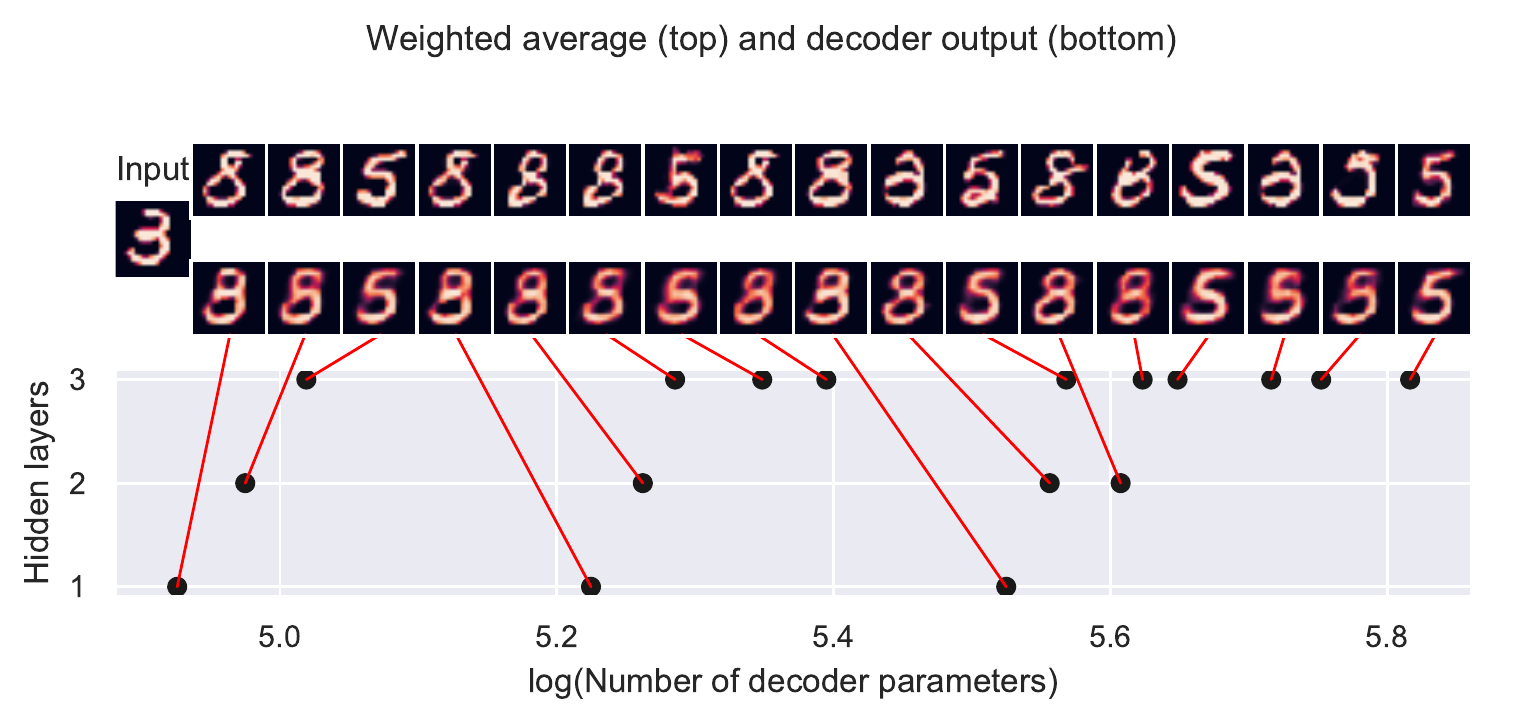}
    \vspace{-0.5em}
    \caption{Decoder outputs and weighted average images for VAEs with different architectures trained on $\text{MNIST}^{tr}_3$.}
    \vspace{-1.0em}
\end{figure}

\begin{figure}[htb]
    \vspace{-1em}
    \centering
    \includegraphics[width=0.9\linewidth]{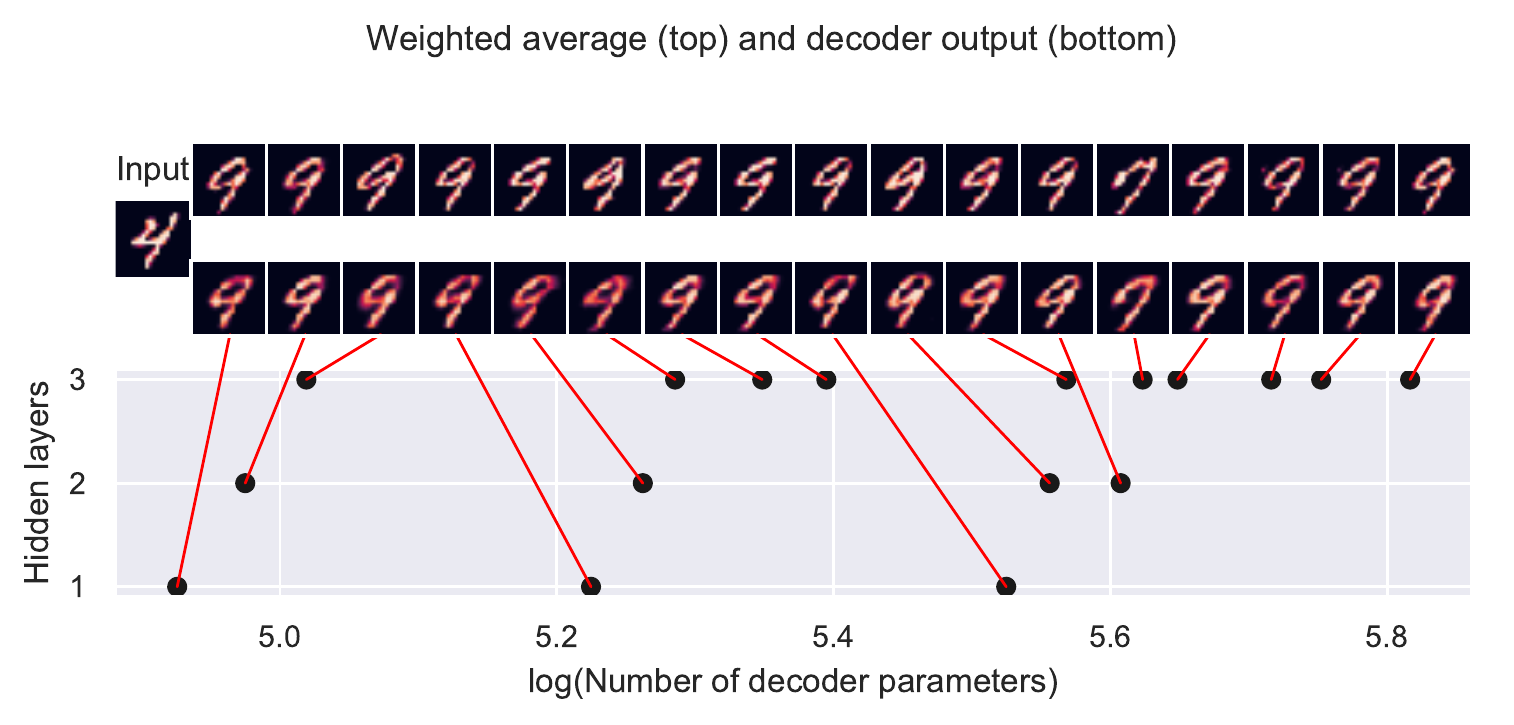}
    \vspace{-0.5em}
    \caption{Decoder outputs and weighted average images for VAEs with different architectures trained on $\text{MNIST}^{tr}_4$.}
    \vspace{-1.0em}
\end{figure}

\begin{figure}[htb]
    \vspace{-1em}
    \centering
    \includegraphics[width=0.9\linewidth]{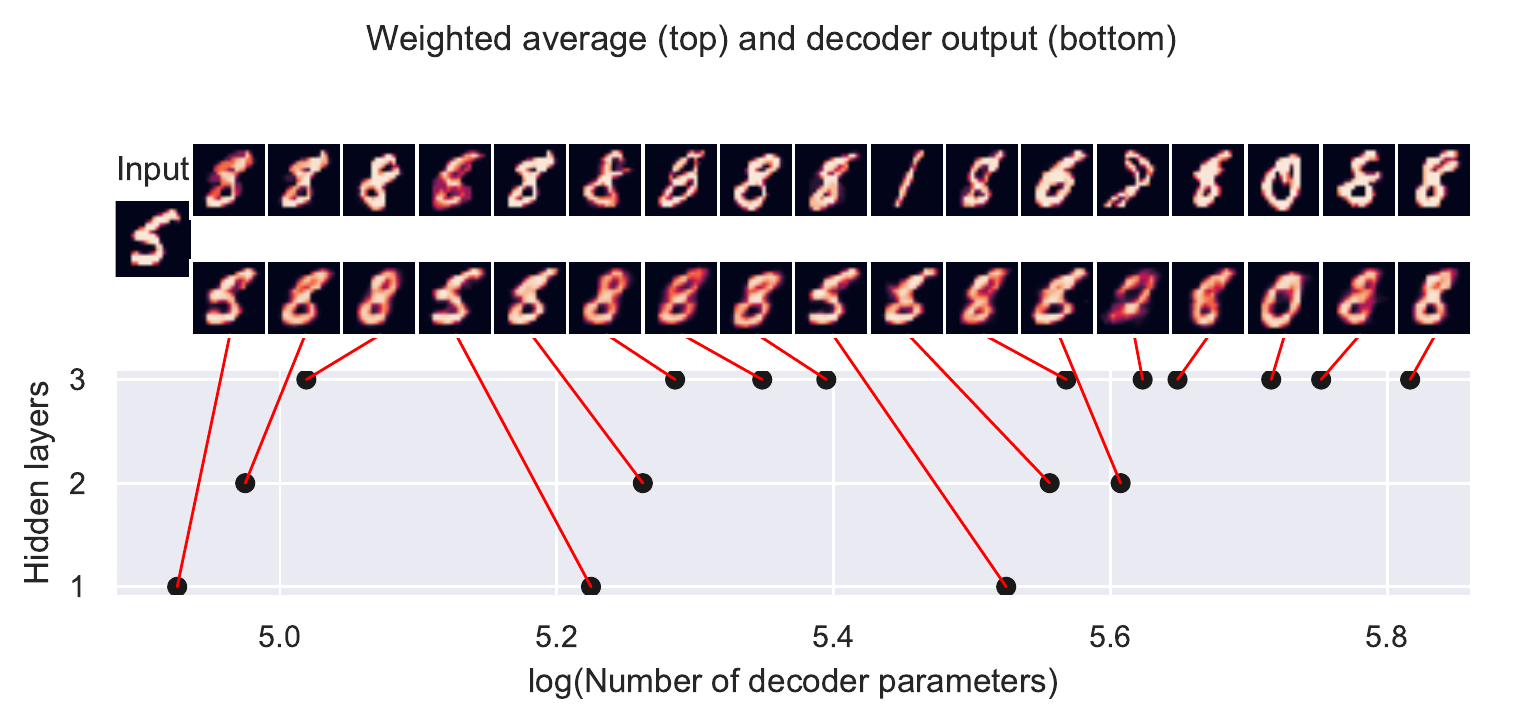}
    \vspace{-0.5em}
    \caption{Decoder outputs and weighted average images for VAEs with different architectures trained on $\text{MNIST}^{tr}_5$.}
    \vspace{-1.0em}
\end{figure}

\begin{figure}[htb]
    \vspace{-1em}
    \centering
    \includegraphics[width=0.9\linewidth]{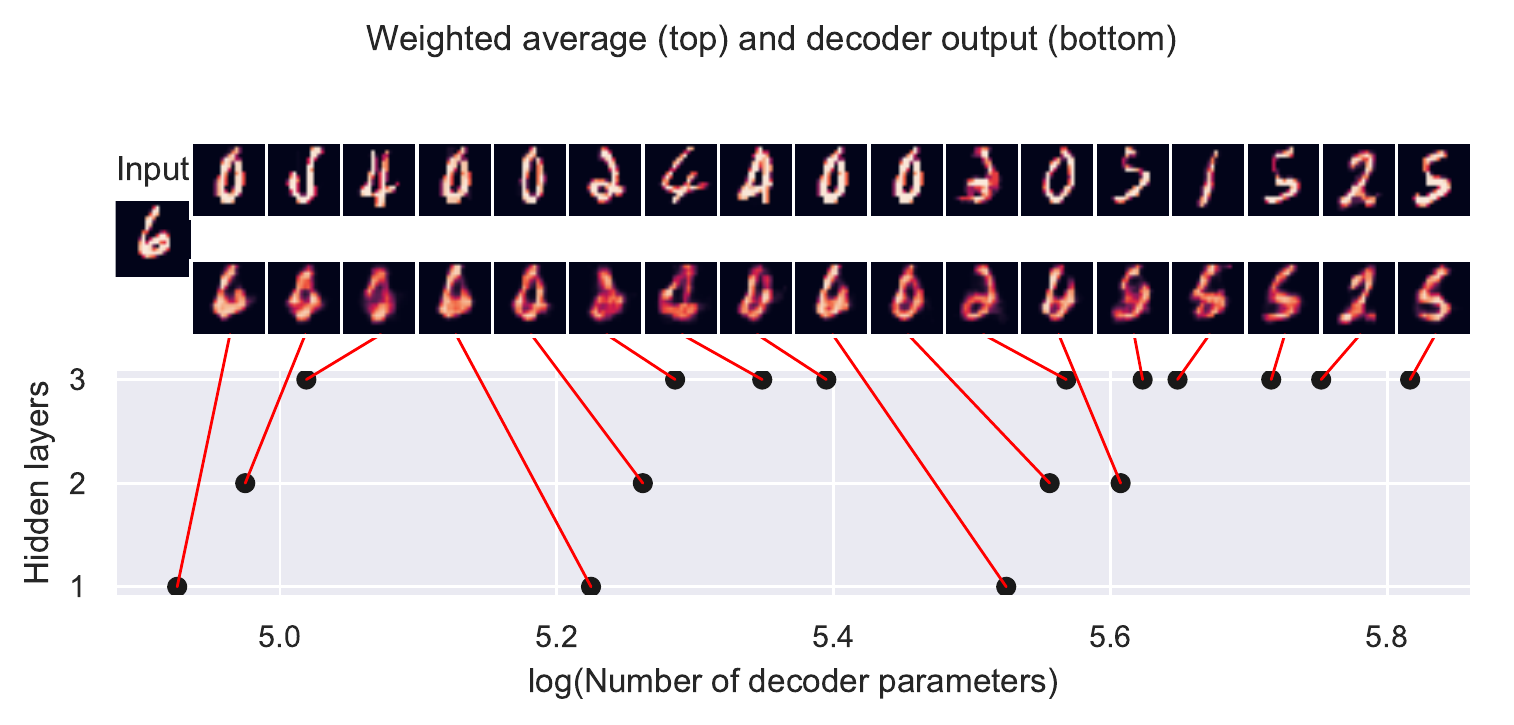}
    \vspace{-0.5em}
    \caption{Decoder outputs and weighted average images for VAEs with different architectures trained on $\text{MNIST}^{tr}_6$.}
    \vspace{-1.0em}
\end{figure}

\begin{figure}[htb]
    \vspace{-1em}
    \centering
    \includegraphics[width=0.9\linewidth]{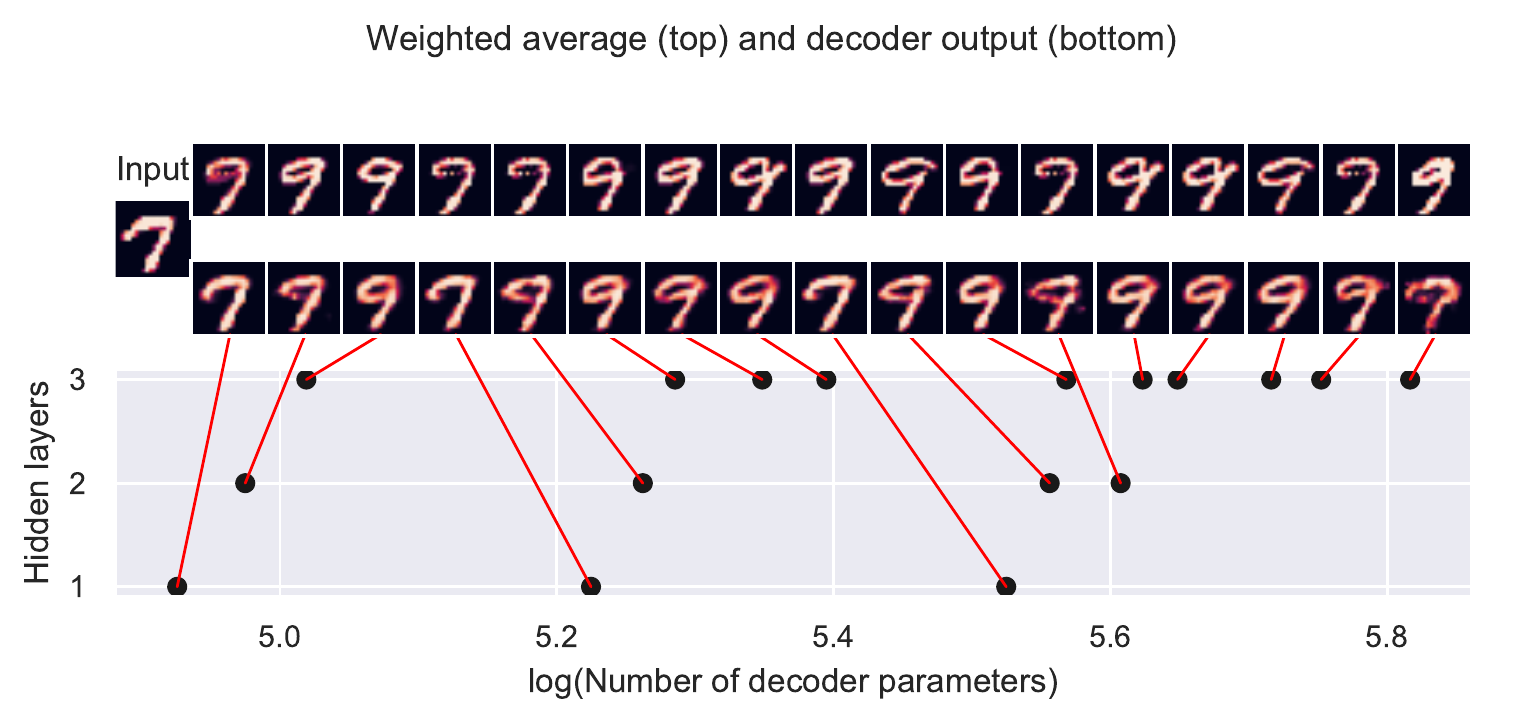}
    \vspace{-0.5em}
    \caption{Decoder outputs and weighted average images for VAEs with different architectures trained on $\text{MNIST}^{tr}_7$.}
    \vspace{-1.0em}
\end{figure}

\begin{figure}[htb]
    \vspace{-1em}
    \centering
    \includegraphics[width=0.9\linewidth]{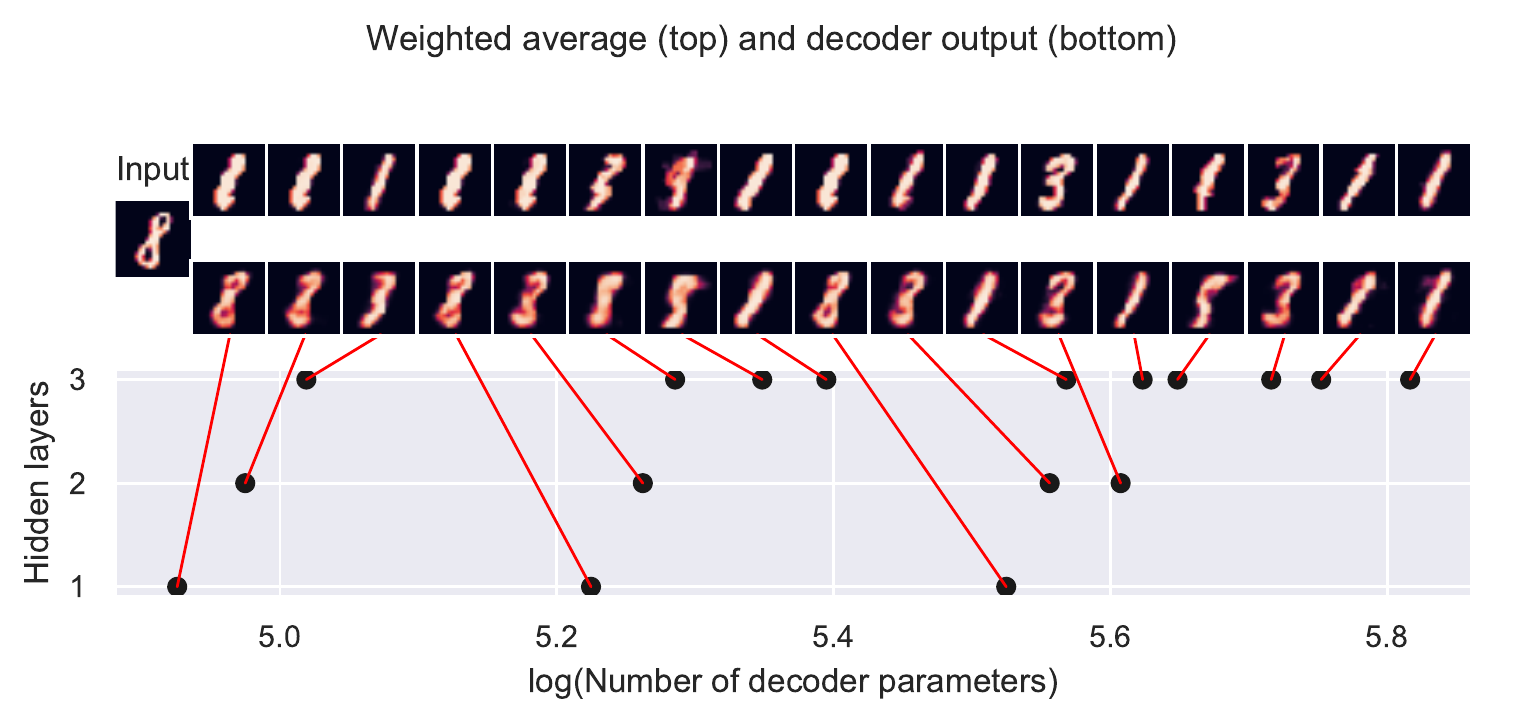}
    \vspace{-0.5em}
    \caption{Decoder outputs and weighted average images for VAEs with different architectures trained on $\text{MNIST}^{tr}_8$.}
    \vspace{-1.0em}
\end{figure}

\end{document}